\documentclass{svproc}
 \usepackage{graphicx} 
 \usepackage{amsmath}
\usepackage[misc,geometry]{ifsym}

\usepackage{url}

\usepackage{orcidlink}

\begin{document}
\mainmatter              % start of a contribution
\title{Towards Edge-Based Idle State Detection in Construction Machinery Using Surveillance Cameras}
\titlerunning{Towards Edge-based Idle State Detection in Construction}  % abbreviated title (for running head)
\author{Xander Küpers \and
Jeroen Klein Brinke\,\orcidlink{0000-0003-4803-2709} \and
Rob Bemthuis\,\textsuperscript{(\Letter)}\,\orcidlink{0000-0003-2791-6070} \and
Ozlem Durmaz Incel\,\orcidlink{0000-0002-6229-7343}}
\authorrunning{X. Küpers, et al.} % abbreviated author list (for running head)
\institute{University of Twente, Enschede, The Netherlands\\
\email{x.l.kupers@student.utwente.nl}\\
\email{\{j.kleinbrinke,r.h.bemthuis,ozlem.durmaz\}@utwente.nl}}

\maketitle              % typeset the title of the contribution

\begin{abstract} %150-200 words
The construction industry faces significant challenges in optimizing equipment utilization, as underused machinery leads to increased operational costs and project delays. Accurate and timely monitoring of equipment activity is therefore key to identifying idle periods and improving overall efficiency. This paper presents the Edge-IMI framework for detecting idle construction machinery, specifically designed for integration with surveillance camera systems. The proposed solution consists of three components: object detection, tracking, and idle state identification, which are tailored for execution on resource-constrained, CPU-based edge computing devices. The performance of Edge-IMI is evaluated using a combined dataset derived from the ACID and MOCS benchmarks. Experimental results confirm that the object detector achieves an F1 score of 71.75\%, indicating robust real-world detection capabilities. The logistic regression-based idle identification module reliably distinguishes between active and idle machinery with minimal false positives. Integrating all three modules, Edge-IMI enables efficient on-site inference, reducing reliance on high-bandwidth cloud services and costly hardware accelerators. We also evaluate the performance of object detection models on Raspberry Pi 5 and an Intel NUC platforms, as example edge computing platforms. We assess the feasibility of real-time processing and the impact of model optimization techniques. 

% We would like to encourage you to list your keywords within
% the abstract section using the \keywords{...} command.
\end{abstract}
\section{Introduction}
\label{section:introduction}
The construction industry is undergoing a significant transformation, driven by increasing demands for more efficient, sustainable, and cost-effective operations~\cite{casini2021construction,musarat2023review}. Challenges such as housing shortages~\cite{algemene-rekenkamer-2022} and stringent environmental regulations underscore the need to optimize construction equipment usage~\cite{rossi2019embedded,zheng2023intelligent}. Construction machinery, which represents a substantial portion of project costs, frequently operates below its optimal efficiency, resulting in unnecessary fuel consumption, elevated emissions, and inflated operational costs. 

The efficient use of machinery, such as excavators, bulldozers and cement mixers, is a major focus to improve project productivity and effective cost management~\cite{kim-2018}. Real-time monitoring and analysis of equipment usage offer a pathway to increased productivity and reduced expenditures~\cite{costin2022iot}. Recent advances in artificial intelligence (AI) and edge computing have supported the development of automated systems that reduce the need for manual oversight, enhance operational workflows, and mitigate environmental impact~\cite{rane2023integrating}. For example,\cite{sepanosian2025iot} demonstrate how IoT-based simulation models can support real-time hazard detection, highlighting the broader potential of distributed sensing and local intelligence on dynamic construction sites. In parallel, real-time IoT architectures have shown promise in monitoring site-level emissions, supporting regulatory compliance and promoting environmental sustainability~\cite{bemthuis2024}. 

Optimizing machinery utilization also leads to significant reductions in fuel consumption and carbon emissions, contributing to both cost savings and environmental sustainability. For instance, a study in~\cite{sizirici-2021} illustrates that increasing the operational efficiency of dump trucks from 40\% to 50\% through the reduction of idle time by six minutes per hour can result in a 10\% decrease in fuel usage and emissions. However, traditional manual monitoring remains labor-intensive, prone to errors, and costly~\cite{azar-2013}. 

Various advanced monitoring systems have been deployed in construction settings, utilizing IoT sensors, drones, RFID, and augmented reality (AR) technologies~\cite{rao-2022}. However, these systems often demand significant computational resources, which limit their adaptability to edge deployment~\cite{azar-2013,chen2020automated,kim-2022}. Many of these systems are server-dependent, leading to increased hardware costs and energy consumption, as well as latency issues caused by continuous data transmission to remote servers~\cite{shi-2016}. 

Existing mobile surveillance camera systems, commonly installed at construction sites for security purposes, present an untapped resource for monitoring machinery usage~\cite{ahmadian2022using,chen2020automated}. By embedding machine learning algorithms into these cameras, they can be reused to track machine efficiency. This integration provides several advantages, such as a reduction in costs by using existing hardware and decreasing bandwidth requirements through edge computing~\cite{liang2024data}. 

This paper proposes a framework, Edge-Based Idle Machine Identification (Edge-IMI), for mobile surveillance cameras to detect idle construction machinery at the edge. The solution uses supervised deep learning, object tracking, and logistic regression techniques, providing real-time insights into machinery utilization. The Edge-IMI framework is composed of the following components: 1) object detection, 2) tracking, and 3) idle state detection. 

Object detection is the process of identifying and localizing objects within images. For the object detection module, we use YOLOv8. Recent case studies suggest notable improvements in detection accuracy and processing speed compared to its predecessors YOLOv3 and YOLOv5~\cite{lee-2024}. This version shows promise for construction applications, with improvements in precision and efficiency highlighted in recent evaluations~\cite{bakirci-2024,sohan-2024}. 

Tracking algorithms are designed to monitor the movement of detected objects over time, ensuring continuous identification even as the object moves across frames. One promising algorithm is ByteTrack~\cite{zhang-2021}, which has shown superior performance in multi-object tracking, particularly in vehicle monitoring applications. ByteTrack surpasses both Simple Online and Realtime Tracking (SORT)~\cite{bewley-2016} and Deep SORT~\cite{wojke-2017} in accuracy, achieving higher object tracking precision, fewer ID switches, and reduced false positives and negatives, all while maintaining fast processing speeds of up to 171 FPS~\cite{abouelyazid-2023}. To our knowledge, while ByteTrack has not yet been widely applied in construction monitoring, its high accuracy and efficiency suggest significant potential for future use in this domain. 

The final component is an idle machinery identification algorithm that determines idle machinery by examining bounding-box data from tracked objects. It assesses the variability in bounding box area and centroid movement during a specified time window and uses a logistic regression model to classify machinery as either idle or active based on this variability. 

We test the performance of the framework on a benchmark dataset created by merging subsets of two publicly available datasets. We demonstrate that the proposed object detection model achieves promising detection performance. We also investigate the computational efficiency of the idle identification algorithm in reliably predicting machinery idle states. We discuss the role of buffer size in managing resource usage, providing insights into possible trade-offs and optimizations. Finally, we evaluate the performance of object detection on a Raspberry Pi 5 and an Intel NUC, focusing on the impact of model optimization techniques. Results indicate that converting the model from PyTorch to OpenVINO significantly improves inference speed. 

The remainder of the paper is structured as follows. Section~\ref{section:literature review} reviews the current literature on idle state identification methods. Section~\ref{section:design and implementation} outlines the design and implementation decisions underlying the proposed Edge-IMI framework. Section~\ref{section:performance evaluation} presents a performance evaluation of the algorithms and Section~\ref{section:discussion} presents a discussion. Finally, Section~\ref{section:conclusion} concludes with a summary of the findings, practical implications, and limitations, followed by an outline of potential future research directions. 

\section{Related Work on Idle Identification Methods}
\label{section:literature review}
Idle state identification is a key component of productivity analysis for construction machinery. One method for detecting idle states involves monitoring changes in the bounding box parameters (e.g., centroid, height, width) over time~\cite{chen2020automated}. This approach uses a sliding window mechanism to compute bounding box statistics, which are then analyzed to determine whether machinery is idle. 

Another approach is the frame difference method, which detects motion based on pixel changes between consecutive frames~\cite{husein-2019}. This method binarizes pixel values, with motion detected when the proportion of white pixels exceeds a pre-defined, adaptive threshold. However, this technique can produce false positives in environments where other objects or workers pass in front of the monitored machinery, making it less suitable for crowded construction sites. 

A third method, which does not rely on visual data, uses on-board sensors to measure parameters such as vehicle speed, location, fuel consumption, and power-take-off status~\cite{yen-2011}. This data can be used to infer idle states, with the vehicle considered idle if it remains stationary for five minutes with the ignition on. While this approach is effective for machinery with embedded sensors, it is not applicable in vision-only systems, such as those using surveillance cameras. Nonetheless, the heuristics from this method can be adapted to vision-based approaches. 

Among these methods, the bounding box-based approach is particularly promising for this study, given its demonstrated effectiveness in the construction domain. Extending this technique to accommodate different machinery types and optimizing it for lower computational costs are key areas of the Edge-IMI framework. Unlike prior methods, the Edge-IMI framework advances bounding box-based idle detection by integrating a logistic regression model designed for low-resource edge computing. This design enhances computational efficiency and enables real-time inference without cloud dependency, offering a scalable and cost-effective solution for construction site monitoring. 

\section{Design and Implementation}
\label{section:design and implementation}
\begin{figure}[ht!]
    \centering
    \includegraphics[width=1.0\linewidth]{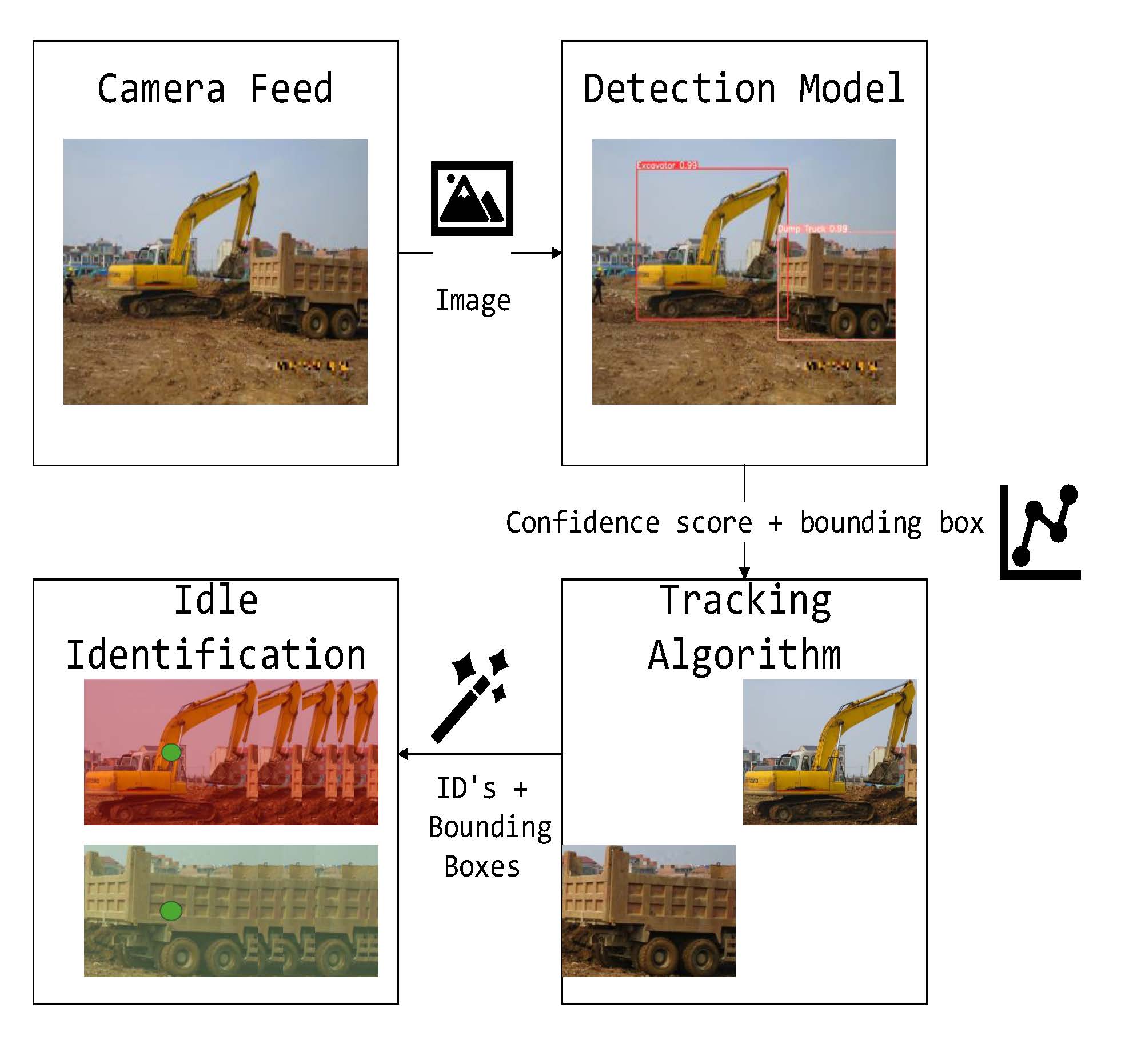}
    \caption{Three components of Edge-IMI framework}
    \label{fig:stages proposed ai framework}
\end{figure}

Fig.~\ref{fig:stages proposed ai framework} illustrates the sequential components of the proposed Edge-IMI framework. In the following, we first introduce the details of the dataset and afterwards focus on each component of the framework. 

\subsection{Dataset Creation}
This study uses data at three distinct stages, each dependent on the output of the preceding stage. The first stage, which involves the detection model, requires an image dataset for training. Several publicly available datasets specifically tailored for machine learning applications in construction were considered. These datasets not only feature construction machinery but also include annotations related to buildings, safety equipment, and construction materials. Among the most comprehensive datasets available are the \textit{Alberta Construction Image Dataset} (ACID)~\cite{xuehui-2021} and the \textit{Moving Objects in Construction Sites} (MOCS) dataset~\cite{xiao-2021}. 

Both datasets provide diverse images captured under various poses, viewpoints, lighting conditions, weather variations, and levels of occlusion, contributing to the model’s robustness under real-world conditions. Combining the ACID and MOCS datasets provides a broader range of environmental conditions, viewpoints, and machinery variations, mitigating the limitations of relying on a single dataset. The basic version of the \textit{ACID} dataset includes three categories of machinery—excavators, dump trucks, and concrete mixer trucks. The \textit{MOCS} dataset contains 13 categories, six of which are relevant to machinery. However, for consistency with the \textit{ACID} dataset, only the overlapping categories—excavators, dump trucks, and concrete mixer trucks—are used. A subset of these two datasets was extracted and merged to create a final dataset. This combination enhances performance metrics by improving generalization across different geographic locations and varied machinery appearances~\cite{shahinfar-2020}. 

\subsection{Data Preprocessing}
While the \textit{MOCS} and \textit{ACID} datasets are pre-labeled, further preprocessing is necessary to tailor them for the detection model. Since only three machinery categories are used from the \textit{MOCS} dataset, irrelevant images and labels are filtered out. Additionally, the \textit{MOCS} dataset is annotated using the Common Objects in Context (COCO) format, which must be converted to the YOLOv8 format. This conversion reduces the annotation information to class ID and bounding box coordinates, meeting the input requirements for YOLOv8. 

Another key preprocessing step involves resizing all images to the YOLOv8-required dimensions of 640x640 pixels, achieved using nearest-neighbor interpolation for computational efficiency. Following this, the bounding boxes and labels are validated using the Computer Vision Annotation Tool (CVAI) to confirm data quality. 

To prevent overfitting, the dataset is split into training and test sets at a 6:4 ratio, with an equal number of images per class—1508 for the training set and 1005 for the test set—ensuring balanced class representation~\cite{ying-2019}. 

\subsection{YOLOV8 Detection Model}
As discussed previously, YOLOv8 is selected as the detection model due to its real-time detection capabilities and efficiency in low-resource environments. YOLOv8 is a single-stage detection architecture, meaning it processes the entire image in a single forward pass, optimizing it for speed while sacrificing some accuracy compared to two-stage models such as Faster R-CNN~\cite{huang-2016}. For this study, YOLOv8 is used without modification, leveraging its pre-trained weights to facilitate transfer learning. The model is trained to detect and classify construction machinery with an inference time of under \textit{150 ms} on a standard CPU. 

The YOLOv8 architecture uses two Convolutional Neural Networks (CNNs) for feature extraction, bounding box regression, and object classification~\cite{sohan-2024}. A pre-trained YOLOv8n model trained on the OpenVINO platform is employed for its smaller size and reduced computational load. Key hyperparameters  were adjusted iteratively to optimize performance as follows: epochs: $100$, batch size: $16$, image size: $640$, optimizer: SGD, learning rate: $0.01$,  momentum: $0.937$ and weight decay: $0.0005$. 

\begin{figure}[ht!]
    \centering
    \includegraphics[width=1.0\linewidth]{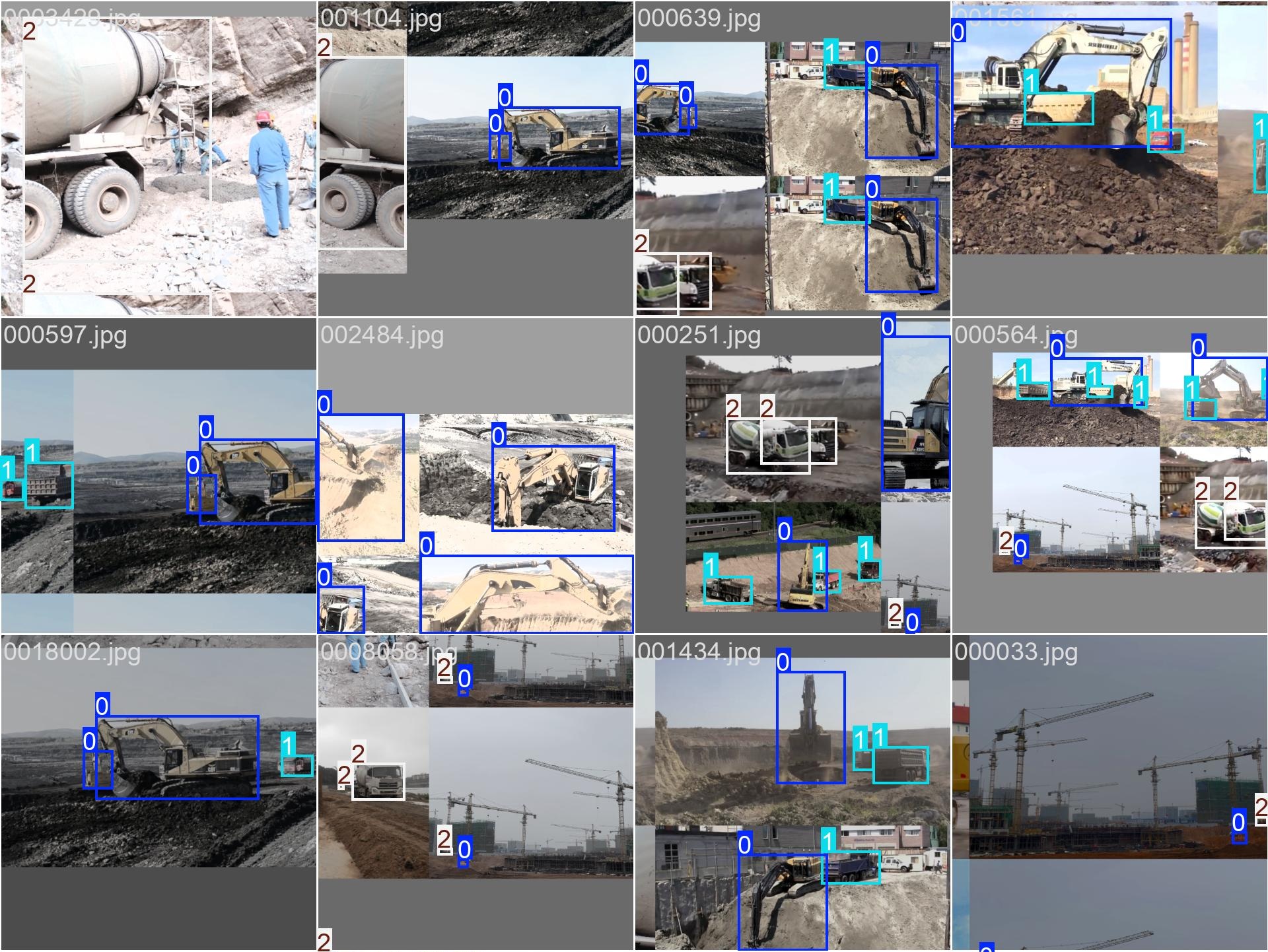}
    \caption{Sample output of the trained YOLOv8 model on validation data}
    \label{fig:sample_output}
\end{figure}

To illustrate the performance of the trained YOLOv8 detection model, Fig.~\ref{fig:sample_output} presents sample predictions from the test dataset. These examples suggest that the model can detect and classify construction machinery across a range of conditions, including occlusions, scale variations, and differing viewpoints. The use of dynamic mosaic data augmentation may have contributed to improved robustness, particularly for less frequent classes such as cement mixer trucks. The results indicate reasonably consistent bounding box placement and class distinction, which could support the model’s applicability in edge-based deployment scenarios. 

\subsection{ByteTrack Tracking Algorithm}
ByteTrack offers advantages over other algorithms in occlusion handling and object identity retention~\cite{zhang-2021}. Its ability to accurately process high- and low-confidence detections with few false positives makes it well-suited for tracking construction machinery in dynamic settings. ByteTrack operates through a track management system that processes the detection output from the model on a per-frame basis~\cite{zhang-2021}. These outputs consist of bounding boxes, confidence scores, class labels, and frame information. Detections are categorized into two groups: high-confidence detections (HCDs), which exceed a high-confidence threshold, or low-confidence detections (LCDs), which fall between the high-confidence and minimal-confidence thresholds. 

For each HCD, the Hungarian algorithm is used to find the best match with existing tracks. If a match is identified, the track is updated with the new detection information. If no match is found, a new track is initialized with a unique track ID. For LCDs, a cost matrix is computed based on the LCD’s features, and the detection is added to the most suitable track. Track maintenance and termination mechanisms are also implemented. A track becomes active after it has been detected in several consecutive frames and is terminated if not updated for a predefined number of frames. Additionally, a re-identification mechanism allows objects reappearing after occlusion to recover their previous track ID, provided their features match sufficiently. 

\begin{figure}[ht!]
    \centering
    \includegraphics[width=1.0\linewidth]{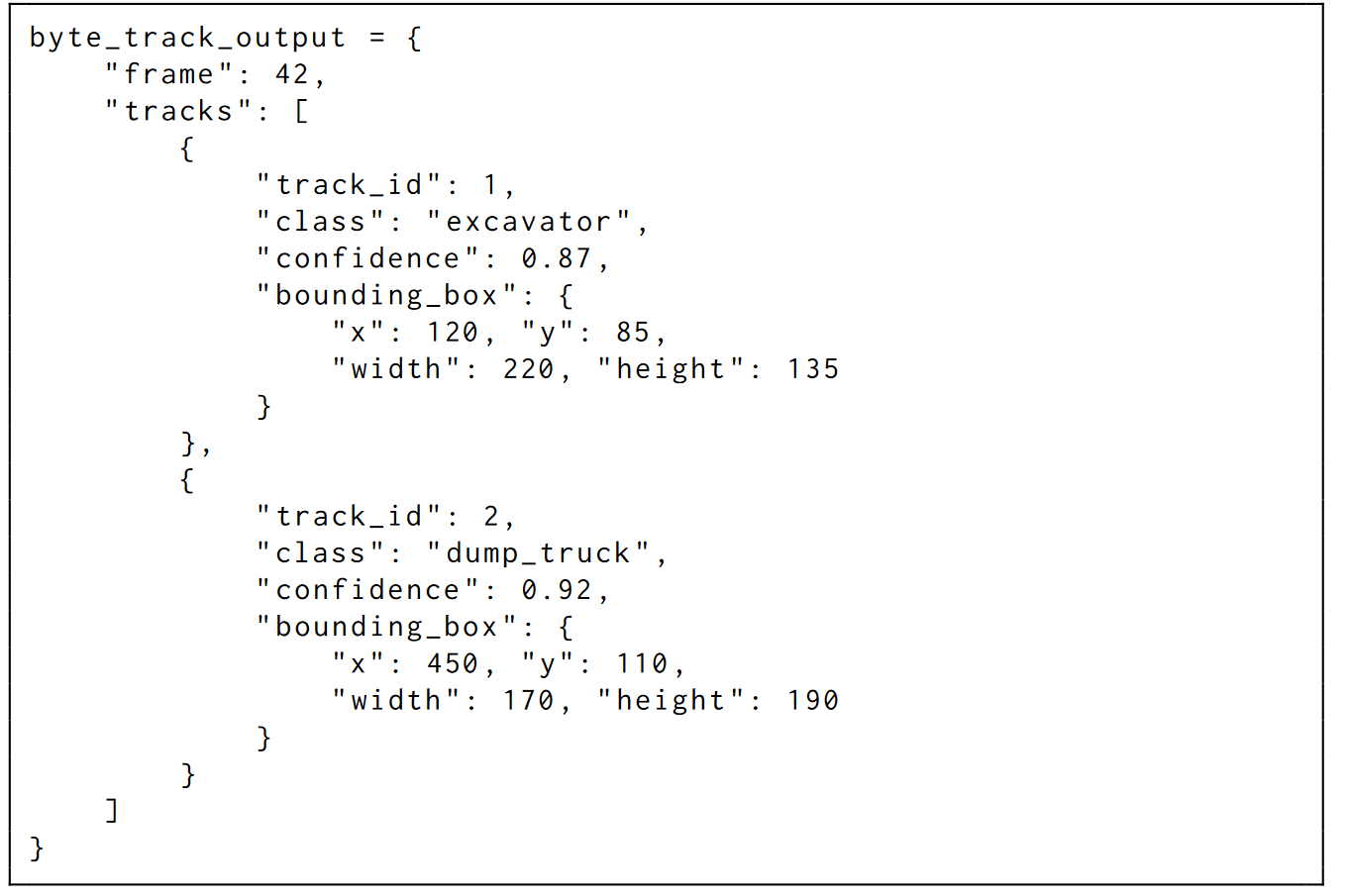}
    \caption{ByteTrack output format}
    \label{fig:byte_track_output}
\end{figure}

No hyperparameter tuning was applied in this study, and the default settings for thresholds, buffers, and minimal bounding box areas were retained, as defined by the YOLO tracker settings for ByteTrack. However, the output format has been customized for this study, as illustrated in Fig.~\ref{fig:byte_track_output}. 

\subsection{Idle Identification Algorithm}
\begin{figure}[ht!]
    \centering
    \includegraphics[width=1.0\textwidth]{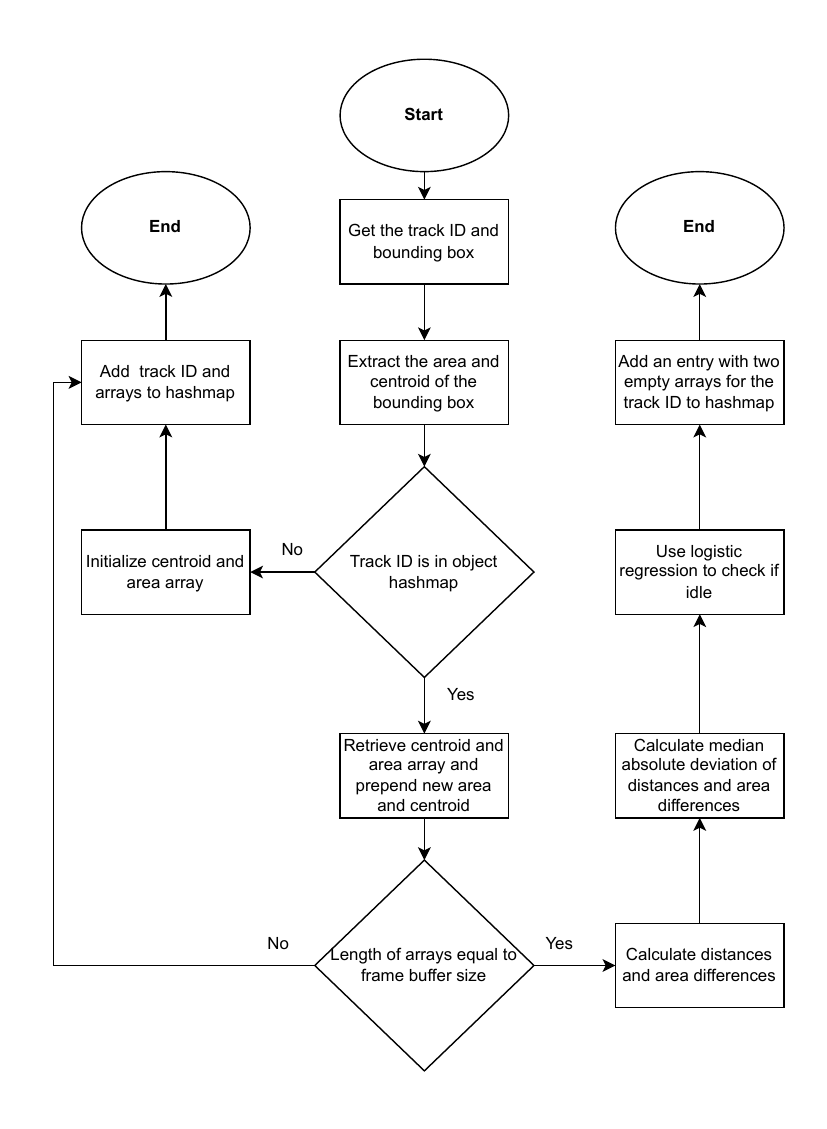}
    \caption{Flow chart of the idle identification algorithm}
    \label{fig:flow chart idle identification}
\end{figure}

The final stage is the identification of idle machinery. We introduce an idle identification algorithm that processes frames using the bounding boxes and track IDs generated by ByteTrack. A hash map is created per frame, with each entry corresponding to a unique Track ID and its associated metadata. An overview of the idle identification process is depicted in Fig.~\ref{fig:flow chart idle identification}. 

Each hash map entry represents a distinct object, identified by its tracking ID. The idle identification algorithm leverages a single feature derived from this metadata—the bounding box—comprising the x and y coordinates of the top-left corner and the bounding box's width and height. This information is sufficient for the idle identification process. 

As each entry is processed by the algorithm, the bounding box’s area and centroid are calculated and stored in two arrays associated with the corresponding tracking ID. A buffer mechanism controls the number of elements in these arrays. Once the buffer is filled, the algorithm performs calculations, after which the buffer is cleared. The buffer size determines how many frames are retained before processing. For example, a buffer size of 20 and a frame rate of 10 frames per second (FPS) results in idle state calculations over a 2-second window, allowing the algorithm to predict whether machinery was idle during that period. We introduce and explain the key parameters of the algorithm in the following.

The Area Difference (AD) given in Eq.~\ref{eq:AD} calculates the change in bounding box area between frames $i$ and $i+1$. The Centroid Difference (CD) given in Eq.~\ref{eq:cd} represents the Euclidean distance between the centroids of the bounding boxes in consecutive frames. The Median Absolute Deviation (MAD) (Eq.~\ref{eq:MAD}), which measures the variability in AD and CD, is used instead of standard deviation due to the non-normal distribution of the data. A Shapiro-Wilk test revealed a p-value of $0.007$, indicating that the data deviates significantly from a normal distribution. MAD is a robust measure of variability, making it suitable for this context. Finally, a logistic regression model (Eq.~\ref{eq:LR}) is applied to classify machinery states as idle or active based on the computed MAD values. If \( p \geq 0.5 \), the machine is classified as active; otherwise, it is deemed idle. The farther p is from 0.5, the greater the model’s confidence in its prediction. 

\begin{equation}
AD = \left| A_i - A_{i+1} \right|
\label{eq:AD}
\end{equation}

\begin{equation}
CD = \sqrt{(x_{i+1} - x_i)^2 + (y_{i+1} - y_i)^2}
\label{eq:cd}
\end{equation}

\begin{equation}
MAD = \frac{\sum_{i=1}^{n} \left| D_i - \text{Median} \right|}{n}
\label{eq:MAD}
\end{equation}

\begin{equation}
p = \frac{1}{1 + e^{-\left( \beta_0 + \beta_1 \cdot \text{MAD\_AD} + \beta_2 \cdot \text{MAD\_CD} \right)}}
\label{eq:LR}
\end{equation}

The logistic regression model was trained on 200 data points using a buffer size of 15 and an FPS of 10, resulting in idle state classifications over a 1.5-second interval. The final model outputs the following coefficients and intercept: \(\beta_0 = 2.4613463131\), \(\beta_1 = -0.00136793\), \(\beta_2 = -0.36581202\). The buffer size and FPS are adjustable parameters that allow customization of the algorithm based on specific objectives. A larger buffer increases the window over which idle predictions are made, but may miss rapid changes in machine activity. Conversely, a smaller buffer improves sensitivity but may result in overreactive predictions. The FPS setting also affects computational load, with higher values requiring more resources but producing more responsive results.

\section{Performance Evaluation}
\label{section:performance evaluation}
The proposed Edge-IMI framework is intended for integration with existing surveillance camera infrastructure. Several experiments were conducted to evaluate its performance across different hardware configurations, including an Intel i3 CPU, 8 GB of RAM, and  no dedicated GPU. The software was tested in a containerized environment, with various metrics used to assess the performance of the detection, tracking, and idle identification models.

\subsection{Evaluation of Object Detection and Tracking Models}
\begin{figure}[ht]
    \centering
    \includegraphics[width=1\linewidth]{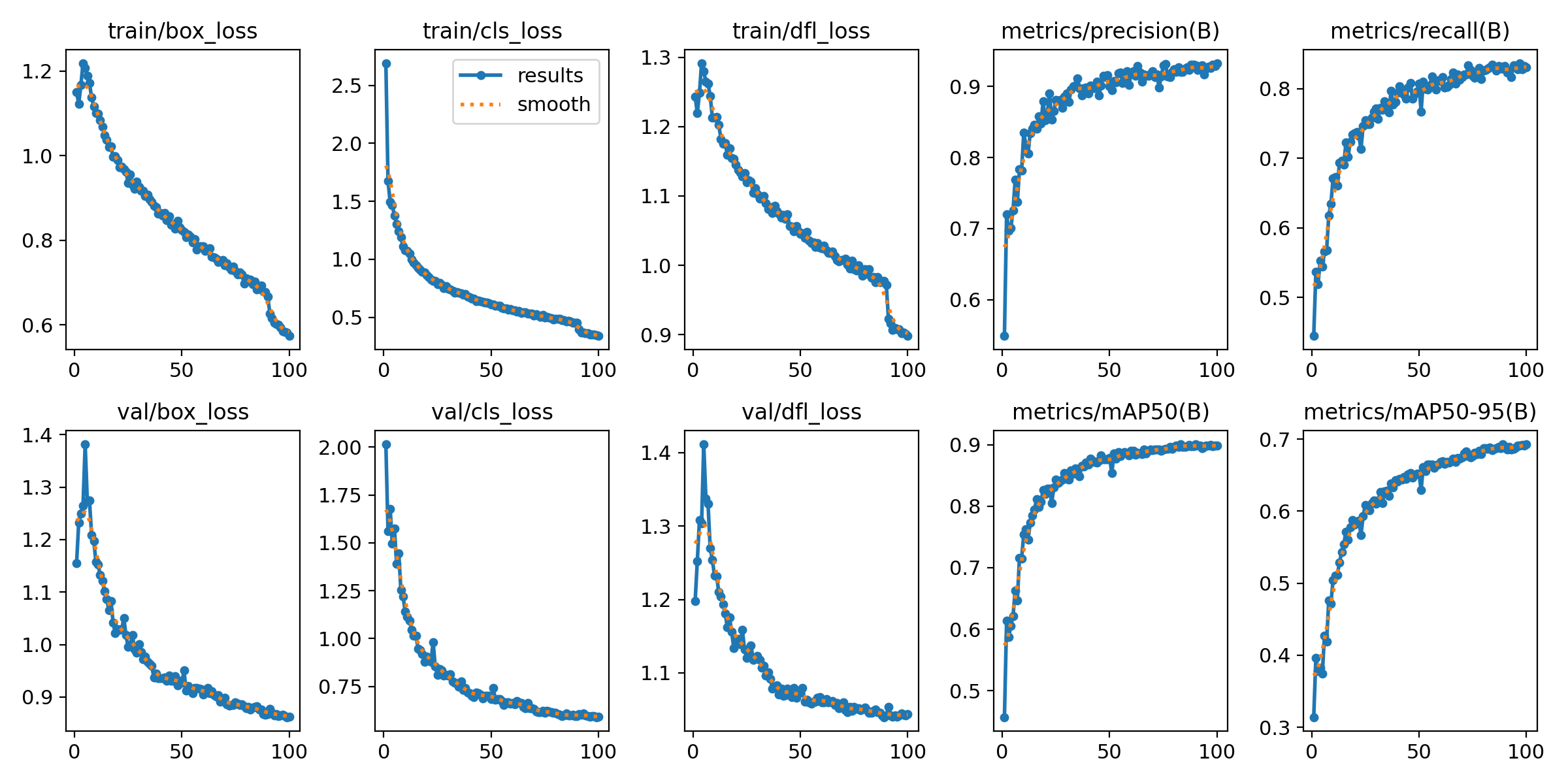}
    \caption{Detection model performance over the course of training}
    \label{fig:Training process model}
\end{figure}

The detection model's performance over $100$ training epochs is illustrated in Fig.~\ref{fig:Training process model}. Table~\ref{tab:evaluation_metrics} summarizes key performance metrics, including precision, recall, $F_1$-score, mAP50, and mAP50-95. In Fig.~\ref{fig:Training process model}, three different types of losses—box loss, class loss, and distribution focal loss—are evaluated for both the training and test datasets. The losses for the test set are higher than for the training set.

\begin{figure}[ht!]
    \centering
    \includegraphics[width=1.0\linewidth]{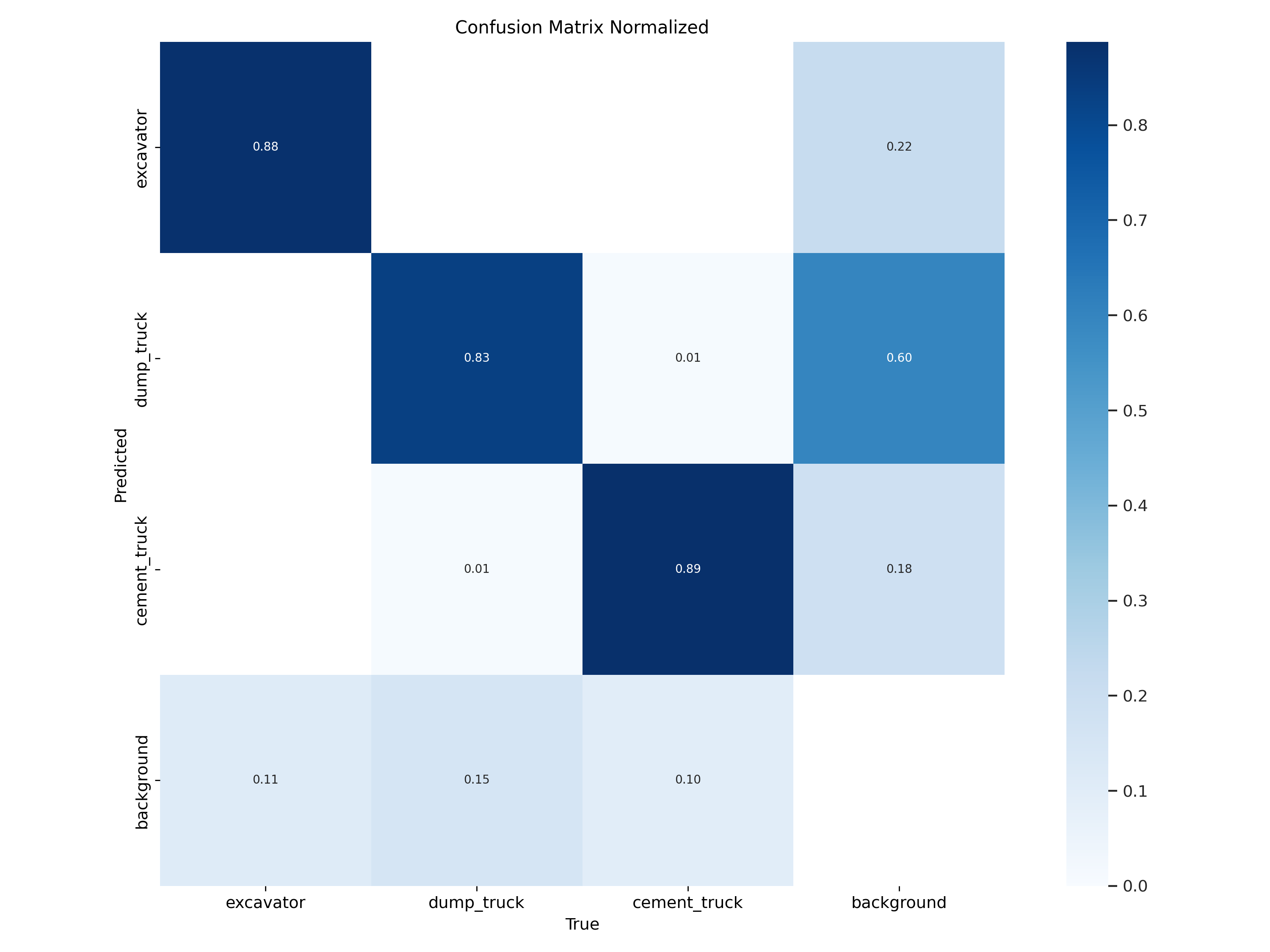}
    \caption{Normalized confusion matrix of the YOLOv8 detection model}
    \label{fig:confusion_matrix}
\end{figure}

To further analyze class-specific performance, Fig.~\ref{fig:confusion_matrix} presents the normalized confusion matrix of the detection model on the test set. The model performs well in classifying excavators and cement trucks, with high true positive rates and limited confusion with other classes. However, dump trucks show considerable misclassification, particularly as background. This may be attributed to visual overlap between dump trucks and common site elements, such as terrain or construction materials. Notably, earlier experiments revealed that the inclusion of background-only images increased the rate of false detections, particularly for dump trucks. As a result, such images were excluded from the final training dataset. These findings suggest that careful curation of background content and more targeted data augmentation may further improve class-specific accuracy, especially for visually ambiguous categories. 

The overall detection model performance is promising ($F_1$-score $=88.00\%$), indicating a balance between precision and recall. The mean Average Precision at 50\% Intersection over Union (mAP50) reached 90.01\%, demonstrating the ability to predict bounding boxes with at least 50\% overlap with ground truth in most cases. However, the lower mAP50-95 suggests variability in bounding box predictions across different Intersection over Union (IoU) thresholds, indicating that fluctuations in the bounding boxes might persist. This could impact idle state detection, requiring threshold tuning in the idle identification algorithm to handle these fluctuations. 

ByteTrack’s performance can be evaluated using metrics like MOTA, MOTP, ID F1 score, ID precision, ID recall, false positives, false negatives, and ID switches. A formal evaluation would require annotated video with bounding boxes, object IDs, and class labels, but due to time constraints, annotation was not performed. However, as the default parameters remain unchanged, the performance is expected to align with the original results~\cite{zhang-2021}.

\subsection{Idle Identification Evaluation}
A test set comprising 150 frames from eight instances was annotated using the CVAI. This annotated data includes the ground truth bounding boxes, class labels, and track IDs. The algorithm’s predictions were then compared against the annotated ground truth, and the accuracy, precision, recall, and F1 score were computed. The results are shown in Table~\ref{tab:evaluation_metrics}. 

The idle identification algorithm demonstrates strong performance, accurately predicting idle machinery states while minimizing false positives. An advantage of the algorithm is its low computational overhead, as it relies primarily on a logistic regression model, making it computationally efficient. However, increasing the buffer size would lead to higher resource consumption. The effect of buffer size on computational overhead has not been rigorously tested in this study, but it is an important factor to consider in future evaluations. 

\begin{table}[ht]
    \caption{Evaluation Metrics}
\centering
    \begin{tabular}{lll}
    \hline
        Metric & Detection & Idle Identification \\
        \hline
        % Precision & 80.08\% & 73.33\%\\
        % Recall & 64.99\% & 79.66\%\\
        $F_1$-score & 71.75\% & 76.42\%\\
        mAP50 & 70.44\% & N/A\\
        mAP50-95 & 55.50\% & N/A\\
        Accuracy & N/A & 78.99\%\\
        %ROC-AUC Score & N/A & 87.71\%  \\
        \hline
    \end{tabular}
    \label{tab:evaluation_metrics}
\end{table}

\subsection{Application Evaluation}
The Edge-IMI framework, tested on construction site surveillance footage, successfully detected excavators and dump trucks, though cement mixer trucks were absent. Excavators, often located at the edges of video frames, had smaller bounding boxes, which reduced centroid and area variation, complicating idle state detection. However, the bounding boxes were still accurately placed for idle identification. The tracking system worked well for dump trucks, with minimal ID switches and smooth transitions, but struggled with excavators, occasionally misidentifying them after extended (10-second) occlusions like tree leaves or sand piles. The idle detection system generally performed well, accurately identifying idle states even for small movements, though there were brief misclassifications when slight motions were undetected. 

\subsection{Deployment and Performance Evaluation of Object Detection on Raspberry Pi and Intel NUC}
We also deployed and evaluated the performance of object detection models on a Raspberry Pi 5 (8GB RAM) and an Intel NUC (i3, 8GB RAM, no GPU) as representative edge devices. Our primary objectives were to assess the feasibility of real-time processing and to evaluate the impact of model optimization techniques—specifically OpenVINO conversion and half-precision (FP16) utilization—on inference speed and accuracy. 

The first evaluation involved processing a 61-second video using the complete Edge-IMI pipeline, which includes object detection, tracking, and idle identification, with a buffer of 15 frames. Performance metrics such as processing time, frames per second (FPS), and mean Average Precision (mAP50-95) were recorded. The results of this end-to-end evaluation are summarized in Table~\ref{tab:Rasp5}. The Raspberry Pi 5 showed limited performance with the original PyTorch model. However, conversion to OpenVINO substantially improved inference speed. Further optimization using FP16 led to a modest increase in FPS but at the cost of an 11\% reduction in mAP50-95, highlighting the trade-off between speed and accuracy. 

\begin{table}[ht]
  \caption{Performance of the complete pipeline, from object detection to idle state identification, on the Raspberry Pi 5}
\begin{tabular}{p{4cm}p{4cm}p{4cm}}
\hline
Model Configuration & Processing Time (s) & Notes  \\ \hline
Original PyTorch Model                             & 244 (excluding initialization)                    & Buffer: 15 frames, Video: 61 seconds \\ 
OpenVINO (Full Precision)                          & 7.18 (61/8.5)                                     & FPS: 8.5                             \\ 
OpenVINO (FP16)                                    & 6.49 (61/9.4)                                     & FPS: 9.4, 11\% mAP drop              \\ \hline
\end{tabular}
\label{tab:Rasp5}
\end{table}

In addition to the full pipeline evaluation, we assessed the inference performance of the YOLOv8 detection-only model using OpenVINO’s benchmarking tools. Table~\ref{tab:vino} reports the inference time and throughput results. On the Intel NUC, the model (12.3 MB, full precision) achieved a throughput of 11.43 FPS, significantly outperforming the Raspberry Pi 5 due to the more powerful CPU. These findings confirm the Intel NUC’s suitability for real-time object detection under constrained edge settings.

\begin{table}[ht]
\caption{Inference time and throughput of the YOLOv8 detection model on the Raspberry Pi 5}
\label{tab:vino}
    \centering
\begin{tabular}{ll}
\hline
Metric & Value \\ \hline
Read Model Time (ms)                  & 18.83                               \\ 
Compile Model Time (ms)               & 337.04                              \\ 
First Inference Time (ms)             & 107.43                              \\ 
Total Execution Time (ms)             & 60184.36                            \\ 
Total Number of Iterations            & 688                                 \\ 
Average Latency (ms)                  & 87.34                               \\ 
Minimum Latency (ms)                  & 78.56                               \\ 
Maximum Latency (ms)                  & 124.28                              \\ 
Throughput (FPS)                      & 11.43                               \\ \hline
\end{tabular}
\end{table}

\section{Discussion and Limitations}
\label{section:discussion}
This study provides project managers and contractors with insights into construction machinery utilization, particularly for evaluating Overall Equipment Effectiveness (OEE)~\cite{singh-2013}. OEE consists of three components: machine uptime (\textit{availability}), the speed and efficiency of operations (\textit{performance}), and the number of tasks that meet required standards (\textit{quality}). The proposed algorithm can supports OEE by detecting idle machinery (availability), tracking machinery and identifying inefficiencies in operations (performance), and preventing overuse or misuse of equipment (quality). Beyond OEE, it also has broader implications, such as cost savings, reduced fuel consumption, and a lower carbon footprint. 

Limited hardware resources in existing surveillance infrastructure, such as low processing power and lack of dedicated GPUs, may hinder real-time performance and idle state detection. While full deployment was not feasible and resource usage in an edge environment was not measured (e.g., model size and inference time), our controlled tests demonstrated strong performance, suggesting potential for real-world application with appropriate privacy measures and resource-efficient strategies. 

While the current results demonstrate feasibility on mid-range edge platforms, such as the Raspberry Pi and Intel NUC, cloud-based deployment may offer higher throughput at the expense of increased latency, potential privacy risks, and network dependency. Edge-IMI is intended to mitigate these trade-offs by enabling fully local inference. Future deployments could extend this approach by distributing pipeline components across heterogeneous edge nodes; for example, executing detection on smart camera modules while delegating tracking and idle classification to nearby compute units. Such a configuration may further reduce latency and improve system efficiency in distributed environments. 

Furthermore, the proposed algorithm has potential limitations that could impact its effectiveness. First, there is a lack of testing in a containerized environment, where resource allocation is shared among containers. This may negatively impact the performance due to limited RAM and CPU resources, especially given the natively lower mAP of a lightweight YOLOv8 model. Additionally, the tracking component relies on static camera input; however, most surveillance systems support pan, tilt, and zoom. These, in turn, disrupt the Kalman filters used in ByteTrack, leading to incorrect track IDs. Lastly, the idle identification component uses a basic logistic regression with limited variables. Testing alternative algorithms and incorporating more features could improve performance. Deployment also posed challenges due to privacy concerns from industry partners, as surveillance footage intended for machinery tracking could inadvertently capture workers, raising data protection issues. Incorporating privacy-preserving methods, such as on-device anonymization or selective masking of human subjects, could further enhance the acceptability of video-based monitoring systems in privacy-sensitive environments. Despite these challenges, continued improvements and broader adoption of Edge-IMI could drive more efficient, cost-effective, and sustainable construction practices. Refinements like advanced idle classification models may further boost accuracy and scalability, making it a robust solution for construction site monitoring. 

\section{Conclusion}
\label{section:conclusion}
This paper presents Edge-IMI, a modular framework for identifying idle construction machinery, designed for deployment on CPU-based edge devices using surveillance camera input. The proposed solution integrates object detection, tracking, and idle state identification, enabling real-time monitoring of machinery utilization using only a CPU. Notably, the YOLOv8 detection model was selected for its balance between accuracy and speed, achieving an F1 score of 71.75\% and an mAP50 of 70.44\%, confirming its reliability in detecting construction machinery. ByteTrack was adopted for object tracking, demonstrating high accuracy and fast processing speeds. Idle state identification is performed using a logistic regression model based on bounding box variability, yielding accurate predictions with minimal computational overhead. The full pipeline was evaluated on a benchmark dataset derived from the ACID and MOCS corpora and tested on representative edge devices, including the Raspberry Pi 5 and Intel NUC. Results confirm the feasibility of real-time inference under constrained hardware conditions, demonstrating the potential of Edge-IMI for integration into existing surveillance systems in resource-limited construction environments. 

While the results are promising, several challenges remain. Future work could focus on integrating video action recognition to assess machinery activities in real-time, although current implementations remain constrained by the need for high-end GPUs. Reducing computational load will be critical for incorporating such capabilities into existing surveillance camera infrastructures. Another promising direction involves enhancing dynamic tracking with support for pan, tilt, and zoom (PTZ) features, along with re-identification of machinery across multiple cameras. This would improve the system’s accuracy in tracking machinery across large or complex sites. Finally, incorporating privacy-preserving techniques and conducting extended real-world deployments will be key steps toward operationalizing Edge-IMI at scale. 

\section*{Acknowledgment}
This work was partly supported by the Dutch Ministry of Infrastructure and Water Management and TKI Dinalog under the ECOLOGIC project (case no. 31192090 and 5000006252). 

\bibliographystyle{splncs03}
\bibliography{samplebib}

\end{document}